\documentclass{article}
\usepackage{ijcai22}

\usepackage{times}
\usepackage{soul}
\usepackage{url}
\usepackage[hidelinks]{hyperref}
\usepackage[utf8]{inputenc}
\usepackage[small]{caption}
\usepackage{graphicx}
\usepackage{stfloats}
\usepackage{amsmath}
\usepackage{amsthm}
\usepackage{booktabs}
\usepackage{algorithm}
\usepackage{algorithmic}
\usepackage{subcaption}
\usepackage{xcolor}

\title{Symbolic image detection using scene and knowledge graphs}

\author{
Nasrin Kalanat\and
Adriana Kovashka\\
\affiliations
Department of Computer Science, University of Pittsburgh\\
\emails
nak168@pitt.edu,
kovashka@cs.pitt.edu
}

\begin{document}

\maketitle

\begin{abstract}
Sometimes the meaning conveyed by images goes beyond the list of objects they contain; instead, images may express a powerful message to affect the viewers' minds. Inferring this message requires reasoning about the relationships between the objects, and general common-sense knowledge about the components. In this paper, we use a scene graph, a graph representation of an image, to capture visual components. In addition, we generate a knowledge graph using facts extracted from ConceptNet to reason about objects and attributes. To detect the symbols, we propose a neural network framework named SKG-Sym. The framework first generates the representations of the scene graph of the image and its knowledge graph using Graph Convolution Network. The framework then fuses the representations and uses an MLP to classify them. We extend the network further to use an attention mechanism which learn the importance of the graph representations. We evaluate our methods on a dataset of advertisements, and compare it with baseline symbolism classification methods (ResNet and VGG). Results show that our methods outperform ResNet in terms of F-score and the attention-based mechanism is competitive with VGG while it has much lower model complexity.
\end{abstract}

\section{Introduction}
\label{sec:Introduction}

\begin{figure}[h]
    \centering
    \includegraphics{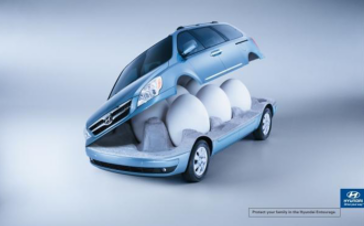}
    \caption{An example of a symbolic image labeled with safety symbol. Eggs symbolically refer to fragility. This helps convey the message of this advertisement image, that this car will protect its passengers, referring to the notion of safety. 
    }
    \label{fig:car_egg}
\end{figure}

Images contained in typical computer vision datasets contain individual objects or complex scenes. Standard prediction tasks defined on these datasets involve object classification or detection, prediction of properties of objects, generating natural-language descriptions of the visual content, etc. However, there are images for which understanding the content of the image extends beyond these prediction tasks. 
In particular, some images \emph{metaphorically} or \emph{symbolically} represent a message, and the information needed to reason about the image lies beyond the list of objects shown inside the image. An example is shown in Figure \ref{fig:car_egg} where a car is metaphorically shown to protect the passengers, by placing eggs, known to be fragile, in place of the passengers.

Both metaphors and symbols are literary devices that use concrete content to make a reference to a more abstract concept. For example, an egg shown in an image can simply be an egg, or it can be a metaphor for fragility.
These intentional visual metaphors are commonly used in marketing, advertisement, politics, and artistic domains for various purposes like increasing influence on the viewers' minds or making the images more attractive. However, detecting these symbolic images and capturing the message they convey is a challenging task. In this study, we try to address the issue.  

To detect symbols, we first need to capture objects, how they relate to each other, and their attributes. For example, in Figure \ref{fig:car_egg} the relation between eggs and car (i.e., eggs ``sit'' in the car) and the feature of eggs (i.e., ``not broken'') are essential to infer the image symbol in this case ( i.e., ``safety''). To capture the visual components, we use a scene graph, a data structure to represent scenes including object instances, their attributes, and the relationships between these objects \cite{johnson2015image}.

Second, we need to reason about common-sense associations between the visual components and abstract concepts.
For instance, in Figure \ref{fig:car_egg} the blue color of the car relates to the concept of calmness and safety. As another example, a red dragon in an image can be a symbol of danger because the red color indicates danger, while a smiling dragon may be a symbol of kindness. Thus, we need a knowledge base to extract the concepts related to the visual components of the image. In this paper, we use ConceptNet \cite{li2016commonsense} which is an open commonsense knowledge base.

We propose a neural network framework, called Scene and Knowledge Graph Symbolic image detection (SKG-Sym) 
, to integrate general knowledge and the visual components of scene graphs in the learning process.  We first construct the scene graph of an input image 
. Then, we extract concepts related to the objects and their features from ConceptNet. Using the extracted concepts, we build the knowledge graph. We use Graph Convolutional Networks (GCN) over each graph to assign weights to visual components and knowledge concepts. GCN is the state-of-the-art method to model graph data. Our intuition behind using GCN is to incorporate the information of the neighborhood of a node in computing its feature representation which capture the relationship network of the image. GCN exploit the information by message passing among the nodes of graphs and compute the representation of each node.
We fuse the outputs of these two steps (scene graph and knowledge graph from ConceptNet) then feed it to a multi-layer perceptron (MLP) to predict the symbols. Afterward, we propose an attention-based version of SKG-Sym, called ASKG-Sym, to improve the proposed framework. We use an attention layer to learn the weights of each graph. 

We evaluate our proposed methods on a dataset of advertisements \cite{hussain2017automatic} 
. The dataset includes following annotations: topic and sentiments, question and answers, symbols, and slogans. Advertisement images metaphorically convey messages which are annotated as symbols in this dataset. As a result, we choose Ads dataset for the evaluation of the methods.
We show that SKG-Sym and ASKG-Sym outperform the baseline method ResNet by a large margin. Moreover, ASKG-Sym achieves the same performance as VGG but needs less than half the number of parameters that VGG needs.

The rest of the paper is as follows: Section \ref{sec:literature-Review} contains related works of prediction tasks on persuasive imagery, and gives an overview of scene graphs and Graph Convolutional Network which are needed for later sections. 
Section \ref{sec:SKG-Sym} introduces our basic framework, SKG-Sym. Section \ref{sec:ASKG-Sym} explains an extension of the framework called ASKG-Sym to improve the approach. Section \ref{sec:Experimental-Results} presents the experimental results obtained on the ads dataset. We conclude the paper in Section \ref{sec:Conclusion}.
\section{Background}
\label{sec:literature-Review}
In this section, we provide a review of the previous work done to predict metaphoric messages. In addition, we explain scene graphs and knowledge graphs because we use them in next sections.

\subsection{Prediction tasks on persuasive imagery}

Joo et al. proposed a method \cite{joo2014visual} to predict the characteristics of a politician using gestures and facial features. In addition, they proposed a method in \cite{joo2015automated} to predict the elections’ outcome using facial features of the candidates.

\cite{hussain2017automatic} introduced an annotated dataset of image advertisements which contains 4 types of annotations such as question and answers and symbols. They showed some classification tasks on this data like question answering and topic and sentiment prediction.
Ye et al. aimed on pairing the correct statement with the corresponding ad \cite{ye2018advise}. They exploited Inception-v4 model to find CNN features of symbol regions. Then they employed an attention approach to aggregate the features of different regions. Moreover, they used triplet loss to make the distance between an image and its corresponding description less than between non-corresponding text and image pairs. 

\cite{guo2021detecting} considered atypicality in an image as a metaphoric expression. To find the atypicality, they used a self-supervised model. They proposed a method that learns the compatibility between objects and context (the rest of the image) using reconstruction losses of masked regions.

\cite{ye2021breaking} proposed a visual reasoning approach which use general knowledge in a generalizable way. They retrieved general knowledge of embedded texts of images using OCR. Then, using the knowledge and objects of the images they build a graph-based model to mask some of the components.

The proposed methods in the area did not consider all visual components including relations between objects in the images and the common-sense knowledge behind the components, which are essential parts of the reasoning process to detect symbols. This paper incorporates visual  components  and  common-sense  knowledge  in  the  symbolic detection process.


\subsection{Scene graphs}
\cite{johnson2015image} first introduced scene graph, a data structure to represent scenes including objects, their attributes and relationships. \cite{krishna2017visual} proposed Visual Genome data set which provides a structured formalized representation of an image. After that, various applications like visual question answering \cite{damodaran2021understanding} and image understanding and reasoning \cite{aditya2018image} used scene graphs
.

\subsection{Knowledge graphs}
We need to incorporate the general knowledge of visual components in the reasoning process of detecting symbols. The most common resources of general knowledge are knowledge bases. A knowledge base is a structured data base which stores facts of the world in an organized way. knowledge base usually is a set of facts. Each fact is a triplet of the form of subject-predicate-object denoting two entities link by a relation. Graphs are a common way to represent the knowledge base where entities are nodes and relations are edges. Google Knowledge Graph, DBpedia, and ConceptNet are examples of such knowledge graphs. In this paper, we use ConceptNet, which is one of the most comprehensive common knowledge graphs \cite{li2016commonsense}. 
It is a multilingual knowledge base built mostly from Open Mind Common Sense \cite{singh2002public}-a system to collect common sense knowledge base from thousands of people online, Games with a purpose to collect common sense \cite{von2006verbosity}, Wiktionary-a free multilingual dictionary, DBpedia- a knowledge base extracted from Wikipedia, and Open Multilingual WordNet- a linked-data representation of WordNet which is a lexical database of semantic relations between words \cite{bond2013linking} 
. 

\section{SKG-Sym}
\label{sec:SKG-Sym}
In this section, we present SKG-Sym framework and its implementation details.
\subsection{SKG-Sym framework}
\begin{figure*}[t]
    \centering
    \includegraphics[width=1\textwidth]{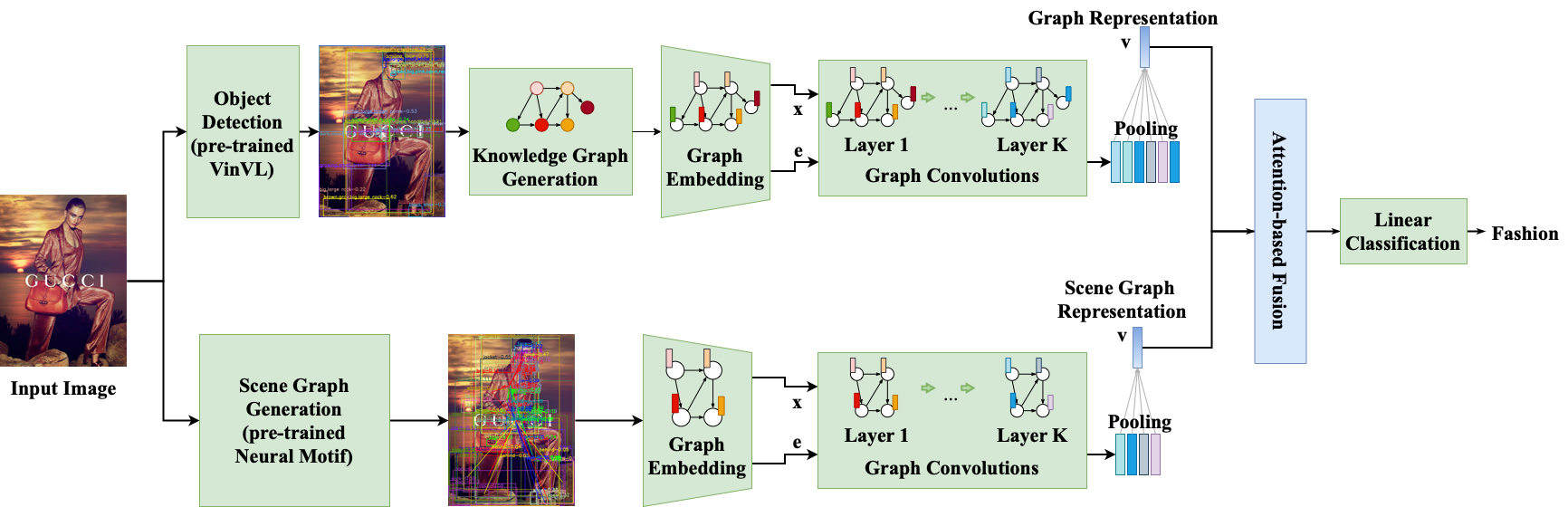}
    \caption{Attention-based Scene and Knowledge Graph for Symbol detection (ASKG-Sym)}
    \label{fig:ASKG-Sym}
\end{figure*}

Our proposed framework, Scene and Knowledge Graph for Symbol detection (SKG-Sym) is the same as the framework shown in Figure \ref{fig:ASKG-Sym} except the fusion component. It consists of seven modules. 




First, in the \textbf{Object Detection} component, we use pre-trained VinVL \cite{zhang2021vinvl} to extract visual information including objects and their attributes. VinVL is pre-trained using following four datasets: including COCO, OpenImagesV5, Objects365V1, and Visual Genome. It keeps 1848 objects, 524 attributes. 

Second, we perform \textbf{Knowledge Graph Generation}. We generate a knowledge graph for each image using the visual information extracted from the Object Detection component. We used ConceptNet \cite{li2016commonsense}
to extract general knowledge for each object and attribute. ConceptNet consists of a set of facts, each represented as a triple of the form $(r, a, b)$, where $r$ is a relation from concept $a$ to concept $b$, e.g. (RelatedTo, car, travel). We consider the extracted concepts as the nodes in the knowledge graph and their corresponding relations as the edges of the graph. 

Third is the \textbf{Scene Graph Generation} component. We use pre-trained Neural motifs \cite{zellers2018neural} to extract scene graphs of images. Neural motifs is pre-trained using Visual Genome. It keeps 150 objects, 50 relations. A scene graph consists of nodes, detected objects in the image, and edges, the relations between objects like belonging to, carrying, between, part of and so on. 

Fourth, we compute a \textbf{Graph Encoding}. We initialize the representation of node $i$ as follows \cite{liang2021graghvqa}:
    \begin{equation}
        v_i^0=\sigma(\frac{1}{|N_i|} \sum_{j\in N(i)}W_{enc}^l [x_j;e_{ij}])
    \end{equation}
where $e_{ij}$ is the word embedding of the relation corresponded to the edge from node j to node i, and $x_j$ is the word embedding of the object and attributes corresponded to node j. 

Fifth, we use \textbf{Graph Convolutional Network} (GCN) to calculate the representation of the graph. GCN uses message passing to model graph data. It uses local neighborhoods information to update the nodes’ features. It calculates the representation of node $i$ in the $l$-th layer $v_i^l$ by averaging the representation of the neighbour nodes $N(i)$ as follows \cite{liang2021graghvqa}:
    \begin{equation}
        v_i^l=\sigma(\frac{1}{|N_i|} \sum_{j\in N(i)}W_{GCN}^l v_j^{l-1})
    \end{equation}
We use the sum function to summarize the representations of nodes from $K$ iterations of message passing $[v_1^K, v_2^K ,..., v_n^K]$ as follows:
    \begin{equation}
     v= Sum([v_1^K, v_2^K ,..., v_n^K])   
    \end{equation}
    
Sixth is a \textbf{fusing} module. We fuse the representation of knowledge graph $v_k$ and scene graph $v_{sg}$ by concatenation and dot product of the vectors as follows:
    \begin{equation}
        v_{fusion}=[v_{kg}, v_{sg}, v_{k}*v_{sg}]
    \end{equation}

Last, we perform \textbf{Classification}: We predict the symbols using an MLP as follows:
\begin{equation}
    Sym=Softmax(MLP(v_{fusion}))
\end{equation}

\subsection{SKG-Sym implementation details}
We consider the facts which are in the set of vocabulary used to train the scene graphs and labels of the images, we consider following relations of knowledge graph: {RelatedTo, IsA, HasA, PartOf, MadeOf, FormOf, AtLocation, Causes, HasProperty, HasFirstSubevent, HasPrerequisite, HasSubevent, UsedFor, CapableOf, DefinedAs, SimilarTo, CausesDesire, Desires, MotivatedByGoal, DerivedFrom}. The relations are the most frequent relations among all 34 relations in ConceptNet.
We used the SGD optimization method to train the model, with a batch size of 32 and learning rate of 1e-3. Word embedding vectors and hidden states are of dimension size of 300 and 512 respectively. We use GloVe for the word embeddings. Our implementation is available publicly on GitHub \footnote{\href{https://github.com/NasrinKalanat/SKG-Sym}{https://github.com/NasrinKalanat/SKG-Sym}}.



\section{ASKG-Sym}
\label{sec:ASKG-Sym}

In the previous section, SKG-Sym naively fuses knowledge graph representation and scene graph representation by concatenation and dot product of the representations. Thus, SKG-Sym consider the graphs equally important. However, the graphs may not be important equally to assign the labels.
To address the issue, we use the attention mechanism to weigh the graphs differently. We compute attention score $\alpha_g$ for the knowledge and scene graphs:



\begin{equation}
    \alpha_{g}=Softmax(v_g^Tv_g)
\end{equation}


We compute attention-based fusion by a weighted average of the representation of the graphs as follows:
\begin{equation}
    v_{Afusion}=\sum_{g \in {kg,sg}}\alpha_g v_g
\end{equation}



\section{Experimental Results}
\label{sec:Experimental-Results}
We evaluate the proposed methods against baselines, experiment ablation study and present quantitative examples in this section.

\subsection{Setup, models and metrics}



Advertisement images metaphorically express messages affecting viewers' minds. 
We evaluate our proposed approaches on the Ads dataset \cite{hussain2017automatic}
, which contains 8,348 ads images paired with 64,131 symbols. Images are multi-labeled. 
Since some of the labels are synonymy and co-occurrence, we cluster the labels into 53 clusters using the method proposed in \cite{hussain2017automatic}.
We use 60\% of the data for training data, 20\% for the validation part and 20\% for the test part.


Based on our knowledge, \cite{hussain2017automatic} is the only work done to predict symbols of Ads data which uses pre-trained ResNet50. In addition, ResNet50 and its competitor VGG16 are the-state-of-the-art frameworks for classification. Thus, we choose them as the baseline methods.
We report F-score as evaluation metric. 

\subsection{Main results}

\begin{figure}[h]
    \centering
    \includegraphics[width=0.5\textwidth]{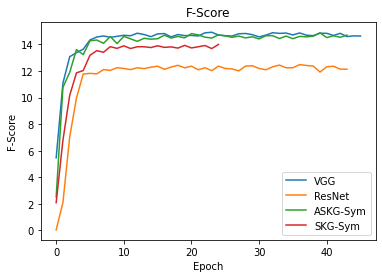}
    \caption{F-score vs epoch number for different methods}
    \label{fig:attention_f}
\end{figure}

Figure \ref{fig:attention_f} shows F-score of each method in each iteration.
We see SKG-Sym outperforms ResNet by a large margin. 
We see ASKG-Sym achieves the same F-score as VGG does. 

Table \ref{tab:model_complexity} shows the number of parameters of each model. We see SKG-Sym works better than ResNet50 even with the lower number of parameters. We see the performance of SKG-Sym is lower than VGG. The reason is that VGG has around six times more number of parameters than VGG has. However, ASKG-Sym achieves comparable results while it needs around half of the number of parameters less than VGG needs to learn.
It is the privilege of our method over VGG.

\begin{table}[h]
\centering
\begin{tabular}{p{2.5cm}ll}
\hline
Method                & Model Complexity & F-score \\ \hline
VGG                            & 138,000,000               & 14.90             \\ 
ResNet50                       & 23,000,000                & 12.50             \\ 
SKG-Sym                        & 17,325,650                & 14.09            \\ 
ASKG-Sym                       & 56,510,088                & 14.86            \\ \hline
\end{tabular}
\caption{Model complexity (number of parameters) and F-score of the methods. 
}
\label{tab:model_complexity}
\end{table}

We show F-score of each label in Table \ref{tab:micro-fscore}. We bold the best number and italicize the second-best number in each row. We see that ASKG-Sym usually gain better F-score for the labels with medium number of frequency.


\begin{table}
\centering
\begin{tabular}{llllll}
\hline
\rotatebox[origin=l]{-90}{label} & \rotatebox[origin=l]{-90}{frequency} & \rotatebox[origin=l]{-90}{SKG-Sym} & \rotatebox[origin=l]{-90}{ASKG-Sym } & \rotatebox[origin=l]{-90}{VGG} & \rotatebox[origin=l]{-90}{Resnet}\\
\hline
fashion & 612 & \textit{28.83} & 28.57 & \textbf{30.51} & 27.31\\
family & 579 & 31.34 & 31.30 & \textbf{38.47} & \textit{34.14}\\
strength & 633 & \textit{36.46} & \textbf{38.02} & 35.08 & 16.29\\
fun & 496 & 0.00 & \textit{1.00} & \textbf{3.36} & 0.00\\
sex & 736 & 9.04 & \textit{12.61} & \textbf{21.92} & 10.92\\
natural & 164 & 0.00 & 0.00 & 0.00 & 0.00\\
comfort & 592 & 26.35 & \textit{28.73} & \textbf{29.26} & 25.00\\
delicious & 234 & 0.00 & 0.00 & 0.00 & 0.00\\
violence & 446 & 0.00 & \textit{2.38} & \textbf{8.83} & 0.00\\
health & 467 & \textit{21.45} & \textbf{22.14} & 19.66 & 21.01\\
speed & 384 & 8.29 & \textbf{9.85} & \textit{8.83} & 8.58\\
energy & 325 & 0.00 & \textit{1.96} & \textbf{9.23} & 0.00\\
love & 351 & 19.53 & 22.90 & \textit{24.03} & \textbf{24.08}\\
entertainment & 368 & 14.96 & \textit{15.45} & \textbf{16.66} & 12.34\\
adventure & 222 & \textit{34.13} & \textbf{36.11} & 27.90 & 9.48\\
vacation & 307 & 20.79 & 19.35 & \textbf{25.00} & \textit{23.07}\\
art & 339 & 16.44 & \textit{16.54} & \textbf{17.61} & 15.44\\
travel & 314 & 22.69 & 20.61 & \textit{24.46} & \textbf{27.43}\\
beauty & 303 & 8.84 & 8.54 & \textit{9.75} & \textbf{10.52}\\
danger & 144 & 0.00 & \textbf{4.59} & \textit{2.17} & 0.00\\
nature & 293 & \textit{9.52} & \textbf{11.11} & 7.65 & 2.08\\
power & 233 & \textit{13.47} & \textbf{16.75} & 9.95 & 4.72\\
sports & 241 & 11.41 & \textbf{16.75} & \textit{13.09} & 11.76\\
happy & 253 & \textit{15.65} & \textbf{18.66} & 13.70 & 14.57\\
freedom & 171 & 0.00 & 0.00 & \textbf{18.18} & \textit{4.34}\\
variety & 249 & \textbf{12.73} & 8.23 & \textit{11.35} & 2.50\\
environment & 252 & 10.47 & 12.50 & \textit{15.00} & \textbf{20.80}\\
technology & 54 & 0.00 & 0.00 & 0.00 & 0.00\\
protection & 212 & 14.20 & \textbf{16.21} & \textit{14.28} & 3.38\\
sexy & 236 & 7.89 & 10.38 & \textit{11.92} & \textbf{16.49}\\
happiness & 222 & \textbf{16.49} & 14.08 & 15.52 & \textit{16.18}\\
party & 178 & \textbf{13.51} & 6.06 & \textit{10.97} & 7.50\\
youth & 196 & 7.63 & \textbf{9.45} & \textit{8.69} & 7.40\\
fitness & 131 & 0.00 & \textbf{3.77} & \textit{3.27} & 0.00\\
safety & 185 & 22.43 & 22.98 & \textit{25.49} & \textbf{32.55}\\
death & 185 & 3.38 & 1.30 & \textbf{11.48} & \textit{4.93}\\
clean & 186 & 7.79 & \textbf{9.95} & 8.88 & \textit{8.98}\\
animal cruelty & 164 & \textit{2.73} & 0.00 & \textbf{8.33} & 0.00\\
relaxation & 192 & \textbf{17.00} & \textit{16.73} & 14.16 & 15.53\\
hot & 172 & \textbf{10.52} & \textit{9.52} & \textit{9.52} & 8.98\\
excitement & 177 & \textbf{7.36} & 3.68 & \textit{4.52} & 4.00\\
healthy & 168 & \textit{16.56} & 15.28 & \textbf{18.68} & 15.74\\
class & 177 & 3.50 & \textit{3.80} & \textbf{8.78} & 2.70\\
christmas & 179 & 7.19 & 7.24 & \textit{9.00} & \textbf{10.32}\\
alcohol & 181 & 8.33 & \textbf{9.39} & \textit{9.27} & 8.19\\
strong & 180 & 13.27 & 12.44 & \textit{16.08} & \textbf{17.50}\\
injury & 186 & \textbf{8.14} & 5.30 & 4.65 & \textit{6.01}\\
hunger & 164 & \textbf{4.76} & \textit{4.56} & 4.44 & 0.00\\
desire & 160 & 11.67 & 9.79 & \textit{14.19} & \textbf{21.87}\\
food & 137 & \textit{20.64} & 20.25 & 19.54 & \textbf{28.23}\\
humor & 169 & 1.79 & \textit{4.87} & \textbf{6.20} & 3.57\\
refreshing & 17 & \textit{2.91} & \textbf{13.07} & 1.47 & 0.00\\
smoking & 140 & 5.69 & 4.14 & \textbf{8.83} & \textit{7.40}\\
\hline
\end{tabular}
\caption{F-score of each label for different methods}
\label{tab:micro-fscore}
\end{table}

\subsection{Ablation}

\begin{figure}[h]
    \centering
    \includegraphics[width=0.5\textwidth]{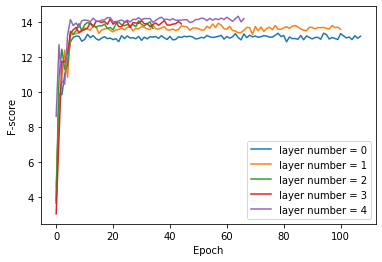}
    \caption{F-score vs epoch number for GCN with different number of layers}
    \label{fig:layers}
\end{figure}

\begin{figure}
    \centering
    \includegraphics[width=0.5\textwidth]{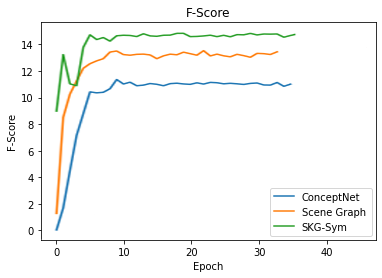}
    \caption{F-score vs epoch number for graph ablation}
    \label{fig:sgconcept}
\end{figure}

To see the impact of the number of GCN layers and each of scene and knowledge graphs, we conduct ablation studies on the layer number (shown in Figure \ref{fig:layers}) and on the graphs (shown in Figure \ref{fig:sgconcept}). 
We find that the more GCN layers improve F-score. The results of 4 layer number are near to 3 layer number, which implies that using more layers would not change the results significantly.

We find that considering both scene graph and knowledge graph leads to achieving a better F-score than using just scene graph or knowledge graph. The performance of using only scene graph is better than using only knowledge graph. The reason is that knowledge graph does not consider the relations of objects in the images. In addition, it considers the concepts related to the objects, some of which may not be important for classifying and may cause the network to be confused.



\subsection{Qualitative examples}

\begin{figure}[t]
     \centering

     \begin{subfigure}[b]{0.23\textwidth}
         \centering
         \includegraphics[width=\textwidth]{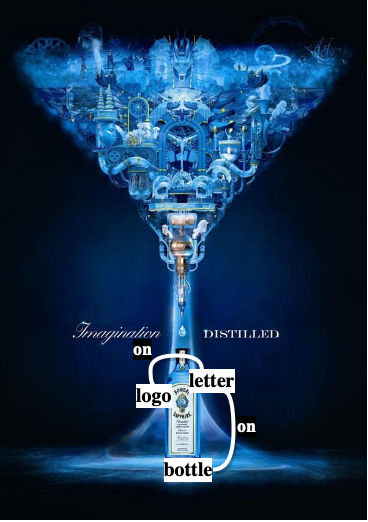}
         \caption{}
         \label{fig:alcohol_sg}
     \end{subfigure}
     \hfill
     \begin{subfigure}[b]{0.24\textwidth}
         \centering
         \includegraphics[width=\textwidth]{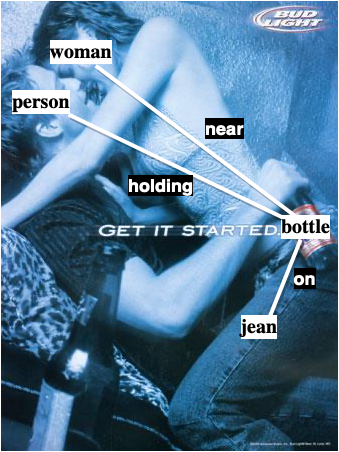}
         \caption{}
         \label{fig:sex_sg}
     \end{subfigure}
     \vfill
     \begin{subfigure}[b]{0.23\textwidth}
         \centering
         \includegraphics[width=\textwidth]{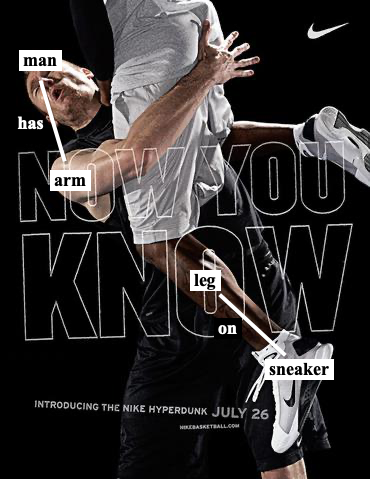}
         \caption{}
         \label{fig:sport}
     \end{subfigure}
     \hfill
     \begin{subfigure}[b]{0.23\textwidth}
         \centering
         \includegraphics[width=\textwidth]{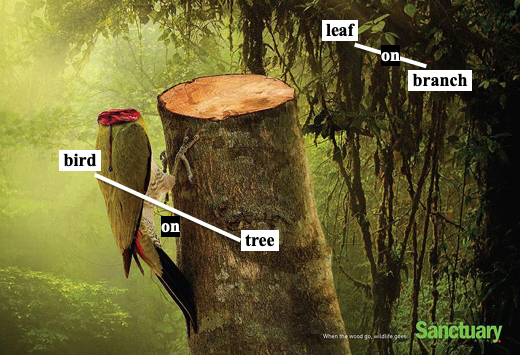}
         \caption{}
         \label{fig:nature}
     \end{subfigure}
    \caption{The most important edges of scene graph for examples of the Ads data with label (a) Alcohol, (b) Sex, (c) Sport, and (d) Nature and Environment}
\end{figure}

We show some examples where our method success in predicting symbols. For each example, we report which one of scene graph or knowledge graph plays an important role.
In addition, we show the edges of scene graph and concepts of knowledge graph that have the highest weights in the network to show the edges and concepts that play silent roles in assigning the label. The example of Figure \ref{fig:alcohol_sg} 
is labeled as alcohol. 
We see, 
the relation between bottle and letter and the relation between logo and letter play important roles. The more weighted concepts of the knowledge graph are as follows:
Bottle IsA a container,
Bottle RelatedTo cup,
Bottle RelatedTo liquid,
Bottle RelatedTo alcohol,
Glass RelatedTo transparent,
Glass RelatedTo crystal,
Glass RelatedTo lime.

It can be observed that ``Bottle RelatedTo alcohol'' 
is among the concepts that play an important role in prediction. In this case, knowledge graph plays a more important role than scene graph.
Figure \ref{fig:sex_sg} shows another example labeled as sex. In this case, the scene graph plays a more important role. The relations between man(person), woman, jeans, and bottle are important in assigning the label Sex.
For \ref{fig:sport}, the scene graph plays important role. In fact, the relations between man, arm, sneaker and leg are important in assigning the label Sport.
For the example in \ref{fig:nature}, knowledge graph plays important role. The most important concepts of knowledge graph for this image are as follows.Bird RelatedTo natural, Animal RelatedTo bird, 
Bird RelatedTo environment, 
Bird AtLocation lawn, 
Plant RelatedTo tree, 
tree RelatedTo nature
. 

\section{Conclusion}
\label{sec:Conclusion}
In this paper, we present a framework to infer symbols of images. The framework utilizes knowledge graph and scene graph. The framework encodes the graphs using multiple iterations of the message through GCN. We evaluate the proposed approaches by comparing it with Resnet and VGG. The results show that our framework works better than Resnet, even with less model complexity. The attention method works like VGG while it needs a much fewer number of parameters, which is our approach's privilege over VGG.
\section*{Acknowledgments}
This research was supported by the Pitt Cyber Accelerator Grants funding.

\bibliographystyle{named}
\bibliography{ijcai22}

\end{document}